\pdfoutput=1

\documentclass[11pt]{article}

\usepackage{acl}

\usepackage{times}
\usepackage{latexsym}

\usepackage[T1]{fontenc}

\usepackage[utf8]{inputenc}

\usepackage{microtype}

\usepackage{inconsolata}

\usepackage{graphicx}

\usepackage{algorithm}
\usepackage{algpseudocode}
\usepackage{multicol}
\usepackage{multirow}
\usepackage{amsmath}
\usepackage{amsfonts}
\usepackage{ulem}
\usepackage{svg}
\usepackage{tabularx}
\usepackage{makecell}
\usepackage{stfloats}
\usepackage{arydshln}
\usepackage{adjustbox}
\usepackage{makecell}
\usepackage{array}
\usepackage{booktabs}
\usepackage{caption}
\usepackage{subcaption}
\usepackage{tikz}
\usepackage{tipa}
\usepackage{pgfplots}
\usepackage{tikz-dependency}
\usepackage{tikz-qtree}
\usepgflibrary{arrows.meta}
\usepgfplotslibrary{fillbetween}

\title{SCOI: Syntax-augmented Coverage-based In-context Example Selection for Machine Translation}

\author{Chenming Tang \quad
Zhixiang Wang \quad
Yunfang Wu\thanks{\ \ \ Corresponding author.} \\
  National Key Laboratory for Multimedia Information Processing, Peking University \\
  MOE Key Laboratory of Computational Linguistics, Peking University\\
  School of Computer Science, Peking University \\
  \texttt{\{tangchenming, ekko\}@stu.pku.edu.cn} \quad
  \texttt{wuyf@pku.edu.cn}
  }

\begin{document}
\maketitle
\begin{abstract}

In-context learning (ICL) greatly improves the performance of large language models (LLMs) on various down-stream tasks, where the improvement highly depends on the quality of demonstrations. In this work, we introduce syntactic knowledge to select better in-context examples for machine translation (MT). We propose a new strategy, namely \textbf{S}yntax-augmented \textbf{CO}verage-based \textbf{I}n-context example selection (SCOI), leveraging the deep syntactic structure beyond conventional word matching. Specifically, we measure the set-level syntactic coverage by computing the coverage of polynomial terms with the help of a simplified tree-to-polynomial algorithm, and lexical coverage using word overlap. Furthermore, we devise an alternate selection approach to combine both coverage measures, taking advantage of syntactic and lexical information. We conduct experiments with two multi-lingual LLMs on six translation directions. Empirical results show that our proposed SCOI obtains the highest average COMET score among all learning-free methods, indicating that combining syntactic and lexical coverage successfully helps to select better in-context examples for MT. Our code is available at \url{https://github.com/JamyDon/SCOI}.

\end{abstract}

\section{Introduction}

In-context learning (ICL) has become a popular prompting strategy to elicit the power of large language models (LLMs) across a wide range of natural language processing (NLP) tasks~\cite{NEURIPS2020_1457c0d6, min-etal-2022-rethinking, dong2023survey}. In ICL, several demonstrations including both task input and ground truth output are presented in the input context, to make LLMs understand the specific down-stream task and produce better results.  

The performance of ICL highly depends on the quality of in-context examples, and it is thus of great significance to explore selecting better examples for ICL~\cite{rubin-etal-2022-learning}. There have been numerous works on in-context example selection for monolingual tasks like natural language inference, commonsense reasoning and semantic parsing~\cite{li-etal-2023-unified, pmlr-v202-ye23c, gupta-etal-2023-coverage,liu2024se2}. Unlike these tasks above, machine translation (MT) involves multiple languages and requires a more sophisticated design of in-context example selection. 
Recently, there have 
some attempts on in-context example selection specially for MT, 
which leverage  
word-level matching~\cite{agrawal-etal-2023-context}, 
embedding-based scoring~\cite{moslem-etal-2023-adaptive, ji-etal-2024-submodular-based, zhu-etal-2024-towards-robust-context} or combination of superficial features~\cite{m-etal-2023-ctqscorer}.

In previous studies, for both
statistical MT and neural MT, syntax plays a crucial  
role in improving model performance~\cite{williams-koehn-2014-syntax, Wu2017ImprovedNM}. However, in case of ICL,
most existing 
works focus on superficial features but pay little attention to the syntactic structure of sentences.  
To achieve a high translation quality, 
it requires not only an accurate word translation but also a proper syntactic structure of the generated target sentence. Hence, syntactic information 
should also play a big part in MT even in the era of LLMs.

Compared with independent selection, it has been proved that selecting in-context examples as an entire set based on the set-level coverage leads to a better diversity 
while reducing redundancy and avoiding sub-optimal results~\cite{gupta-etal-2023-coverage}. As a typical NLP task, 
MT would also benefit from in-context examples with a high set-level coverage. 
Therefore, beyond the conventional 
lexical coverage, 
high syntactic coverage is also necessary to select 
informative in-context examples for MT. 


In this work, we propose \textbf{S}yntax-augmented \textbf{CO}verage-based \textbf{I}n-context example selection, SCOI~\footnote{/\textipa{'skoUI}/.}, to boost LLMs' performance on MT. Specifically, 
to measure syntactic coverage, we first simplify a tree-to-polynomial algorithm~\cite{liu2022quantifying}, which is 
originally costly but has been reduced to no more than quadratic time complexity after simplification.
Using this 
new algorithm, we convert syntax trees into polynomials and then compute the set-level syntactic coverage based on 
vector representations of polynomial terms. 
Meanwhile, we 
compute the proportion of word overlap to measure set-level lexical coverage. After that, we design an alternate approach to combine both coverage measures, so that word-level and syntax-level features would complement each other.

We evaluate SCOI on 6 translation directions (German, French, Russian into and out of English) based on two open-source multi-lingual LLMs, XGLM$_\text{7.5B}$~\cite{lin-etal-2022-shot} and Alpaca~\cite{alpaca}. Among all learning-free methods, SCOI obtains the highest COMET scores on 4 out of 6 translation directions and the highest average COMET score. Especially, on Russian-to-English and English-to-Russian translations, SCOI even outperforms the learning-based CTQ Scorer~\cite{m-etal-2023-ctqscorer} when using Alpaca.

Our contributions can be summarized as follows:
\begin{itemize}
    \item Going beyond superficial word matching, we introduce the knowledge of syntactic structure  
    to in-context example selection for MT.
    \item To take advantage of both word overlap and syntactic resemblance, we propose a novel framework
    to ensure a high set coverage at both word and syntax level for in-context example selection, and empirical experiments validate the effectiveness of our method.
    \item We 
    design a simplified tree-to-polynomial algorithm 
    owning a complexity upper bound of no more than quadratic time.
    In contrast, that of the original version could be polynomial time with an arbitrarily large degree.

\end{itemize}

\section{Related Work}
Prompting LLMs for better performance has been one of the mainstream trends of NLP research. There have been a large number of studies on prompting strategies for MT in recent years~\cite{vilar-etal-2023-prompting,pmlr-v202-zhang23m}. \citet{puduppully-etal-2023-decomt} decompose the translation process into a sequence of word chunk translations to improve LLMs' performance on translation between linguistically related languages. \citet{ghazvininejad2023dictionarybased} propose to present LLMs with a set of possible translations for a subset of the input words from bilingual dictionaries to improve LLMs' performance on low-resource and out-of-domain MT. \citet{10.1162/tacl_a_00642} prompt LLMs with selected knowledge including keyword pairs, topics and sentence pairs to emulate human-like translation. \citet{zhang2024teaching} manage to teach LLMs an unseen language on the fly with the help of a small parallel corpus and a dictionary. \citet{guo-etal-2024-teaching-large} first create a textbook including vocabulary list, language examples with syntax patterns and translate instructions using LLMs and then prompt LLMs with the textbook just created to better translate low-resource languages. \citet{zhu-etal-2024-towards-robust-context} prompt LLMs with both sentence-level and word-level demonstrations, the former selected with a margin-based score and the latter being word pairs most related to the test input appeared in the former.

Among various prompting strategies, ICL plays a key role. \citet{rubin-etal-2022-learning} suggest that the performance of ICL strongly depends on the selected in-context examples. Thus it is of great significance to select better examples using various strategies. \citet{li-etal-2023-unified} propose to train a unified demonstration retriever for ICL across a wide range of tasks. \citet{pmlr-v202-ye23c} make use of determinantal point processes (DPPs) to ensure both relevance and diversity of examples. \citet{liu2024se2} select examples in a sequential rather than "select then organize" way that leverages the LLM's feedback on varying context, aiding in capturing inter-relationships and sequential information among examples. \citet{gupta-etal-2023-coverage} define measure of set-level information coverage and select examples based on it, which inspires our work.

There are some example selection strategies customized for MT. \citet{agrawal-etal-2023-context} select examples based on n-gram overlap. \citet{moslem-etal-2023-adaptive} select examples based on sentence embedding similarity. \citet{m-etal-2023-ctqscorer} train language-specific regression models to combine various features for example selection. \citet{ji-etal-2024-submodular-based} select examples based on submodular functions combining surface/semantic similarity and diversity within examples. To the best of our knowledge, no previous work has made use of syntactic information in in-context example selection for MT.

\section{Method}
\label{sec:method}
\begin{figure*}[htbp]
    \centering

    \begin{adjustbox}{center}
    \includegraphics[scale=0.49]{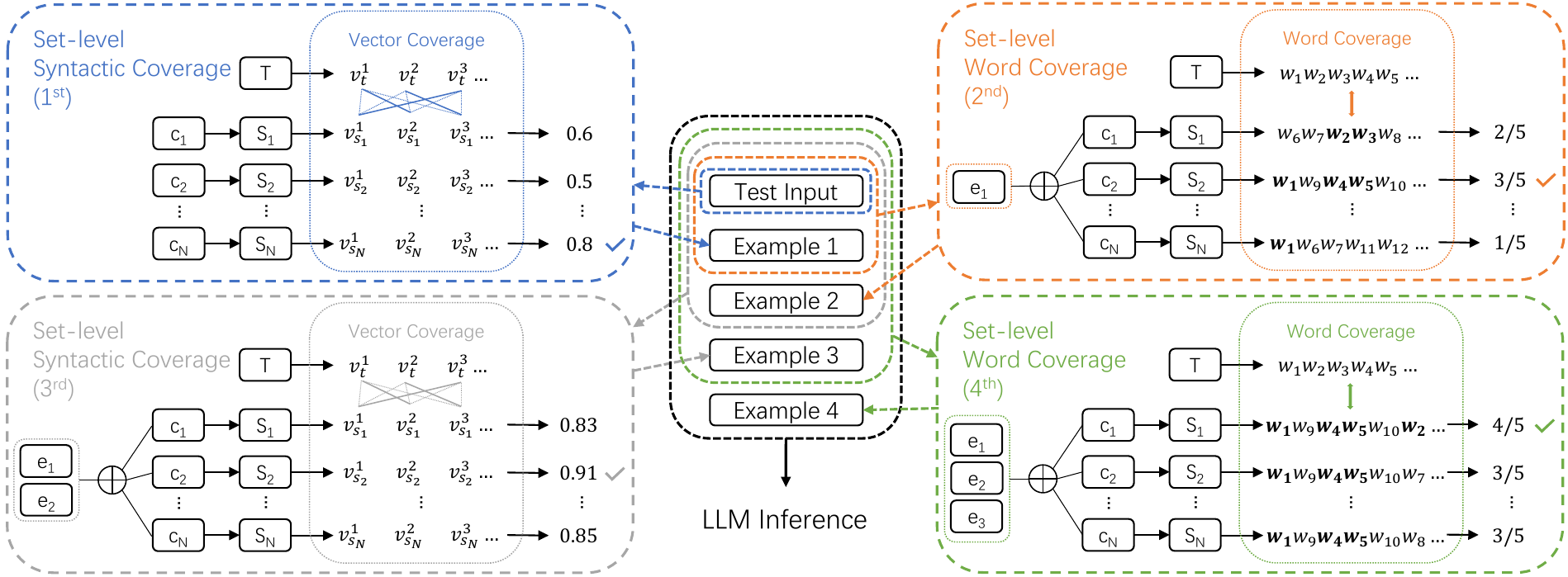}

    \end{adjustbox}
    
    \caption{Overview of SCOI. 
    Each example is selected based on how well the test input is covered by the current candidate plus the existing examples selected in previous steps at syntax level and word level alternately. In each step, $T$, $e_i$, $\oplus$, $c_i$, $S_i$ denote the test input, the $i$-th selected example, concatenation of selected examples and one candidate, the $i$-th candidate from the example database, the to-be-scored set including the selected examples plus the $i$-th candidate, respectively.}
    \label{fig:overview}
\end{figure*}

We propose to select in-context examples based on both syntactic and lexical coverage to better apply LLMs for MT. 
Specifically, to measure the set-level syntactic coverage, we first simplify a tree-to-polynomial algorithm, making it practical to run on large MT datasets, and then compute the coverage of vector representations of polynomial terms. 
To measure the set-level lexical coverage, we simply
consider the proportion of word overlap. After that, we design an alternate strategy to take advatage of both lexical and syntactic knowledge. An overview of our proposed method, SCOI, is presented in Figure~\ref{fig:overview}.

\subsection{Polynomial Representation of Syntactic Structure}
\citet{liu2022quantifying} convert dependency trees into polynomials recursively and compute the distance between polynomials to 
measure the syntactic similarity between sentences from different languages. Specifically, given the number of dependency labels $d$, dependency trees will be transformed into polynomials based on two variable sets: $X=\{x_1, x_2...x_d\}$ and $Y=\{y_1, y_2, ...y_d\}$. Considering a leaf node with label $l$ as $n^l$, its corresponding polynomial is $P(n^l, X, Y) = x_l$. For a non-leaf node $m^l$ with label $l$, its polynomial is:
\begin{equation}
\label{eq:original}
    P(m^l, X, Y) = y_l + \prod_{i=1}^{k}P(n_i, X, Y),
\end{equation}
where $n_1, ..., n_k$ are all child nodes of $m^l$.

However, the algorithm can be of very high complexity when the dependency tree is large. 
In MT, there 
are often millions of data to be processed and it is thus impractical to make use of the original algorithm from \citet{liu2022quantifying}. 
Therefore, 
we propose a simplified polynomial algorithm,  reducing the complexity of tree-to-polynomial conversion to no more than quadratic time. 

Concretely, our newly defined polynomial is based on only one variable set $X=\{x_1, x_2...x_d\}$. For a leaf node $n^l$, its polynomial remains $P(n^l, X) = x_l$. For a non-leaf node $m^l$ with child nodes $n_1, ..., n_k$, its polynomial is:
\begin{equation}
\label{eq:simplified}
P(m^l, X) = x_l\cdot (1 + \sum_{i=1}^{k}P(n_i, X)).
\end{equation}
where each term $x_1^{e_{x_1}}x_2^{e_{x_2}}...x_d^{e_{x_d}}$ in the polynomial corresponds to a path from the root node to one certain node in the tree, and $e_{x_i}$ indicates the number of nodes with the $i$-th dependency label on that path. 
Given a sentence with a dependency tree rooted in Node $r$, the polynomial representing the syntactic structure of that sentence is $P(r, X)$.

We analyze the complexity of the original and our simplified tree-to-polynomial algorithms in Appendix~\ref{app:analysis}.

\subsection{Measure of Set-level Syntactic Coverage}
\label{subsec:syncov}
Given a test input $x$ and a set of in-context examples $Z$, the set of salient aspects (e.g., entities, keywords, etc.) of $x$ being $S_{x}$, 
the set-level information coverage of in-context examples is defined as \cite{gupta-etal-2023-coverage}:
\begin{equation}
\label{eq:setcov}
    \mathtt{SetCov}(x, Z) = \sum_{s\in S_{x}}{\mathop{\max}\limits_{z\in Z}\mathtt{c}(s, z)},
\end{equation}
where $\mathtt{c}(s, z)$ measures the coverage or recall of a single salient aspect $s$ by example $z$.

For better parallelization and to better fit the salient aspects denoting syntax in this work, which are vector representations of polynomial terms from the tree-to-polynomial algorithm, we reformulate Equation~\ref{eq:setcov} to the 
set-level syntactic coverage:
\begin{equation}
\label{eq:synsetcov}
    \mathtt{SynSetCov}(x, Z) = \frac{1}{\vert T_x \vert}\sum_{s\in T_{x}}{\mathop{\max}\limits_{t\in T_Z}\mathtt{c}(s, t)},
\end{equation}
where $T_{x}$ is the multiset~\footnote{Since we take repeated elements into account, we use \textit{multiset}~\cite{hickman1980note} that allows repetition of elements instead of \textit{set} in this work.} of terms in the polynomial representation of the dependency tree of $x$, $T_Z = \bigcup_{z\in Z}{T_z}$ is the multiset of all the terms in polynomials of dependency trees of all the in-context examples in $Z$, $s$ and $t$ denote terms in $T_{x}$ and $T_Z$ respectively, and $\mathtt{c}(s, t)$ computes the similarity of term $s$ and term $t$.

To compute $\mathtt{c}(s, t)$, we first compute the distance between the two polynomial terms. 
Note that a term $t = x_1^{e_{x_1}}x_2^{e_{x_2}}...x_d^{e_{x_d}}$ can be written as a term vector with $d$ entries:
\begin{equation}
\label{eq:vec}
    v_t = [e_{x_1}, e_{x_2}, ..., e_{x_d}],
\end{equation}
where each entry represents the exponent of the corresponding variable. The distance between terms $s$ and $t$ can thus be 
calculated by the Manhattan distance \cite{Craw2017} between vectors $v_s$ and $v_t$:
\begin{equation}
\label{eq:manhattan}
    \mathtt{d}(s, t) = {\parallel v_s - v_t \parallel}_1.
\end{equation}

As distance is negatively correlated with similarity, 
we compute $\mathtt{c}(s, t)$ using the normalized distance:
\begin{equation}
\label{eq:norm}
    \mathtt{c}(s, t) = \frac{1}{1 + \mathtt{d}(s, t)}.
\end{equation}
In this way, $\mathtt{c}(s, t)$ is a normalized value between $0$ and $1$. Note that each term in the polynomial represents a root-to-node path in the tree. So $\mathtt{SynSetCov}(x, Z)$ indicates the average coverage of each path in the dependency tree of $x$ by all the dependency trees in $Z$.

\subsection{Measure of Set-level Lexical Coverage}
\label{subsec:lexcov}
In this work, we simply measure the set-level lexical coverage by computing the proportion of word overlap:
\begin{equation}
    \mathtt{WordSetCov}(x, Z) = \frac{\vert W_x \cap W_Z\vert}{\vert W_x\vert},
\end{equation}
where $W_x$ is the multiset of the words in $x$ and $W_Z = \bigcup_{z\in Z}{W_z}$ is the multiset of all the words in all the examples in $Z$.

\subsection{Combining Syntactic and Lexical Coverage}
\label{subsec:combcov}
Combining syntax-level and word-level coverage could make them complement each other and thus help select better in-context examples for MT. In this work, we propose an alternate way to combine both.

For convenience, we number ICL examples starting from 1. Specifically, for each odd-numbered
example, we select it based on how well the current candidate, along with the existing examples selected in previous steps, covers the test input in syntax, while for each even-numbered example, we select it based on set-level lexical coverage. To put it more concretely,
we select the first example with the highest set-level (only the first example at this time) syntactic coverage and the second example with the highest set-level (including the first and the second example) lexical coverage.

\begin{algorithm}[tb]
    \scriptsize
    \caption{Greedy Optimization of Set Coverage} %
    \label{alg:cov}
    \begin{algorithmic}[1]
        \Require Example database $\mathcal{T}$; test input $x$; desired number of demonstrations $k$; coverage scoring function $\mathtt{SynSetCov}$ and $\mathtt{WordSetCov}$.
        \State $Z \gets \emptyset$\Comment{Selected in-context examples.}
        \State $Z_{\text{curr}} \gets \emptyset$\Comment{Current set cover.}
        \State $\mathtt{curr\_syn\_cov} \gets -\inf$
        \State $\mathtt{curr\_word\_cov} \gets -\inf$
        \While{$|Z| < k$}
            \If{$|Z| \equiv 0 \pmod{2}$}
            \Comment{Odd-numbered to-be-selected example.}
                \State $z^*, \mathtt{next\_syn\_cov} = \operatornamewithlimits{argmax}\limits_{z \in \mathcal{T}-Z} \mathtt{SynSetCov} \left(x, Z_{\text{curr}} \cup z\right)$
                \If{$\mathtt{next\_syn\_cov} > \mathtt{curr\_syn\_cov}$}
                \Comment{Pick $z^*$.}
                    \State $\mathtt{curr\_syn\_cov} \gets \mathtt{next\_syn\_cov}$
                    \State $Z \gets Z \cup z^*$
                    \State $Z_{\text{curr}} \gets Z_{\text{curr}} \cup z^*$
                \Else
                \Comment{Start a new one if no increase.}
                    \State $Z_{\text{curr}} \gets \emptyset$, ~~$\mathtt{curr\_syn\_cov} \gets -\inf$
                \EndIf
            \Else
            \Comment{Even-numbered to-be-selected example.}
                \State $z^*, \mathtt{next\_word\_cov} = \operatornamewithlimits{argmax}\limits_{z \in \mathcal{T}-Z} \mathtt{WordSetCov} \left(x, Z_{\text{curr}} \cup z\right)$
                \If{$\mathtt{next\_word\_cov} > \mathtt{curr\_word\_cov}$}
                \Comment{Pick $z^*$.}
                    \State $\mathtt{curr\_word\_cov} \gets \mathtt{next\_word\_cov}$
                    \State $Z \gets Z \cup z^*$
                    \State $Z_{\text{curr}} \gets Z_{\text{curr}} \cup z^*$
                \Else
                \Comment{Start a new one if no increase.}
                    \State $Z_{\text{curr}} \gets \emptyset$, ~~$\mathtt{curr\_word\_cov} \gets -\inf$
                \EndIf
            \EndIf
        \EndWhile
        \State \textbf{return} $Z$
    \end{algorithmic}
\end{algorithm}

Following \citet{gupta-etal-2023-coverage}, we use a greedy algorithm to select the optimal set as shown in Algorithm~\ref{alg:cov}. It alternately selects examples that lead to the maximum syntactic coverage (lines 7-11) and lexical coverage (lines 16-20). If no example brings further increase in coverage, the algorithm reserves the selected examples and starts another round (lines 12-13 and 21-22).

\section{Experimental Setup}
We follow \citet{m-etal-2023-ctqscorer} to set up our experiments.
\subsection{Datasets and Evaluation Metrics}
\begin{table}[htbp]
\small
  \centering
    \begin{tabular}{cccc}
    \toprule
    \textbf{Language} & \textbf{ISO Code} & \textbf{Dataset} & \textbf{\#Pairs (M)} \\
    \midrule
    German & DE    & Europarl & 1.83 \\
    French & FR    & Europarl & 1.92 \\
    Russian & RU    & ParaCrawl & 5.38 \\
    \bottomrule
    \end{tabular}
  \caption{Data statistics.}
  \label{tab:data}
\end{table}

\paragraph{Test Set} We perform our evaluation on the \textit{devtest} set of FLORES-101~\cite{goyal-etal-2022-flores}, which has 1012 sentences with translations in 101 languages. We experiment between English and 3 common languages including German, French and Russian.
\paragraph{Example Database}
We use Europarl~\cite{koehn-2005-europarl} for German and French and ParaCrawl~\cite{banon-etal-2020-paracrawl} for Russian as example database. Detailed statistics are shown in Table~\ref{tab:data}.
\paragraph{Evaluation Metrics}
We report COMET~\cite{rei-etal-2020-unbabels} scores from \texttt{wmt20-comet-da}, which is considered a superior metric for MT nowadays~\cite{kocmi-etal-2021-ship}. We report BLEU scores from sacreBLEU~\cite{post-2018-call} in Appendix~\ref{app:bleu}.

\subsection{Pre-processing}
We parse all the datasets with spaCy~\cite{Honnibal_spaCy_Industrial-strength_Natural_2020} to get dependency trees for our syntax-based approaches. The spaCy models we use for different languages are listed in Appendix~\ref{app:spacy}.

We use Sacremoses~\footnote{https://github.com/hplt-project/sacremoses} to tokenize all the languages for the lexical coverage computation.

\subsection{Large Language Models}
XGLM$_\text{7.5B}$~\cite{lin-etal-2022-shot} and Alpaca~\cite{alpaca} are used in our experiments. XGLM is a multilingual generative language model supporting 30 languages and has 7.5B parameters in total. Alpaca is a 7B model fine-tuned from LLaMA~\cite{touvron2023llama} on 52K instruction-following data.

\subsection{Implementation Details}
The number of in-context examples is set to 4 in our experiments. 

For XGLM, we use the same prompt template as used in \citet{m-etal-2023-ctqscorer}:

\begin{verbatim}
    [source] sentence: [X_1]
    [target] sentence: [Y_1]
    ###
    ...
    [source] sentence: [X_k]
    [target] sentence: [Y_k]
    ###
    [source] sentence: [X]
    [target] sentence:
\end{verbatim}

In the template, \texttt{[source]} and \texttt{[target]} refer to the names of the source and target languages in English (e.g., German, French, etc.). The \texttt{\#\#\#} symbol is used as an example delimiter and a marker for answer extraction in post-processing.

With the same symbols above, for Alpaca, we use the same template as used in \citet{10.1162/tacl_a_00642}:

\begin{verbatim}
Instruction: Translate the following
[source] text into [target].

[source]: [X_1]
[target]: [Y_1]
...
[source]: [X_k]
[target]: [Y_k]
[source]: [X]
[target]:
\end{verbatim}

Noting that our test data and example databases are the same as those used in \citet{m-etal-2023-ctqscorer}, we directly use the examples selected by BM25, R-BM25 and CTQ Scorer from \citet{m-etal-2023-ctqscorer}~\footnote{https://github.com/AI4Bharat/CTQScorer}.

Following \citet{m-etal-2023-ctqscorer}, we remove instances in the example database with more than 120 tokens in order to avoid overlong context.


\subsection{Baselines}
\label{subsec:baseline}
\paragraph{Zero-shot:}
No in-context examples are provided.

\paragraph{Random:}
Examples are selected randomly for each test input from the example database. We report the average result of 3 different random seeds.

\paragraph{BM25:}
We use the BM25 algorithm implemented by \citet{Bassani_retriv_A_Python_2023} to retrieve the top-$k$ matching examples in the example database for each test input.

Following \citet{agrawal-etal-2023-context} and \citet{m-etal-2023-ctqscorer}, all the compared methods below re-rank examples based on top-100 examples retrieved by BM25 for each test input.

\paragraph{R-BM25:}
We evaluate R-BM25~\cite{agrawal-etal-2023-context} for comparison, which ensures n-gram coverage and diversity.

\paragraph{Fuzzy:}
We evaluate Fuzzy~\cite{moslem-etal-2023-adaptive}, where examples that are most similar in sentence-level embedding are selected. We use sentence transformers~\cite{reimers-2019-sentence-bert} with \texttt{paraphrase-multilingual-MiniLM-L12-v2} \cite{reimers-2020-multilingual-sentence-bert} to reimplement it.

\paragraph{CTQ Scorer:}
We evaluate CTQ Scorer~\cite{m-etal-2023-ctqscorer} for comparison, which is a learning-based method combining multiple features including number of tokens, similarity in LaBSE embeddings~\cite{feng-etal-2022-language}, perplexity, etc. It trains a specific regression model for each language pair.

\paragraph{SCOI:}
Our proposed method described in Section~\ref{sec:method}.

\section{Results and Analysis}
\subsection{Main Results}
\begin{table*}[htbp]
\small
  \centering
    \begin{tabular}{cccccccc}
    \toprule
    \multirow{2}[4]{*}{\textbf{System}} & \multicolumn{3}{c}{\textbf{Into EN}} & \multicolumn{3}{c}{\textbf{Out of EN}} & \multirow{2}[4]{*}{\textbf{Avg.}} \\
\cmidrule(lr){2-4}\cmidrule(lr){5-7}          & \textbf{DE} & \textbf{FR} & \textbf{RU} & \textbf{DE} & \textbf{FR} & \textbf{RU} &  \\
    \midrule
    \midrule
    \multicolumn{8}{c}{\textbf{XGLM}} \\
    \midrule
    Zero-shot & 60.26 & 70.40 & 50.63 & -28.39 & -5.13 & -123.67 & 4.02 \\
    \midrule
    \multicolumn{8}{l}{\textit{Learning-free}} \\
    Random & 63.53 & 70.80 & 53.41 & 43.03 & 53.23 & 42.70 & 54.45 \\
    BM25  & 63.21 & 71.36 & 52.48 & 44.13 & \textbf{55.54} & 44.58 & 55.22 \\
    R-BM25 & 64.13 & 71.18 & 54.06 & 44.83 & 55.21 & 45.92 & 55.89 \\
    Fuzzy & 64.40 & \textbf{71.92} & 53.37 & 44.45 & 55.23 & 44.69 & 55.68 \\
    SCOI (\textit{ours}) & \textbf{64.67} & 71.26 & \textbf{54.08} & \textbf{44.87} & 55.31 & \textbf{46.47} & \textbf{56.11} \\
    \midrule
    \multicolumn{8}{l}{\textit{Learning-based}} \\
    CTQ Scorer & 65.38 & 70.65 & 53.48 & 45.52 & 56.00 & 48.59 & 56.60 \\
    \midrule
    \midrule
    \multicolumn{8}{c}{\textbf{Alpaca}} \\
    \midrule
    Zero-shot & 68.95 & 76.12 & 57.13 & 41.01 & 54.41 & 24.66 & 53.71 \\
    \midrule
    \multicolumn{8}{l}{\textit{Learning-free}} \\
    Random & 69.71 & 76.64 & 57.47 & 42.60 & 56.58 & 28.61 & 55.27 \\
    BM25  & 69.08 & 76.41 & 58.52 & 43.65 & 57.34 & 32.63 & 56.27 \\
    R-BM25 & 69.71 & \textbf{76.70} & 57.69 & 43.87 & \textbf{59.17} & 34.78 & 56.99 \\
    Fuzzy & 69.72 & 76.36 & 58.12 & \textbf{44.10} & 57.25 & 30.57 & 56.02 \\
    SCOI (\textit{ours}) & \textbf{69.79} & 76.08 & \textbf{58.66} & \textbf{44.10} & 57.97 & \textbf{36.26} & \textbf{57.14} \\
    \midrule
    \multicolumn{8}{l}{\textit{Learning-based}} \\
    CTQ Scorer & 70.39 & 76.57 & 58.63 & 45.55 & 58.71 & 35.68 & 57.59 \\
    \bottomrule
    \end{tabular}
  \caption{COMET scores of 4-shot ICL performance of SCOI and other methods for translation on all 6 directions of 2 language models. The zero-shot baseline of each model is listed in the first row. All methods except CTQ Scorer are learning-free, which do not require task, language or LLM-specific training. "\textbf{Avg.}" refers to the average score across all 6 directions. The highest scores among learning-free methods are in \textbf{bold} text.}
  \label{tab:main-comet}
\end{table*}

Main results are shown in Table~\ref{tab:main-comet}. SCOI obtains the highest COMET scores of 4 out of 6 translation directions and the highest average COMET score among all learning-free methods using both XGLM and Alpaca, showing competitive performance across language models. Using XGLM, SCOI outperforms the learning-based CTQ Scorer on "RU-EN", while using Alpaca, SCOI even outperforms CTQ Scorer on both "RU-EN" and "EN-RU". Note that Alpaca seems not good at generating Russian, and its performance gain with 4-shot random examples is fairly poor compared with the zero-shot baseline. But SCOI greatly improves its performance on "EN-RU" and shows amazing ability in teaching an LLM to better translate into a language that appeared less during training.

We observe that SCOI shows obvious preference across different languages. For example, it fails to benefit "FR-EN" but improves performance on "RU-EN". Besides different natures of different languages, this might be also due to different capabilities of syntax parsers for different languages. We find that when the parser performs poorly (e.g., the French parser), SCOI also performs less competitively, while a more powerful parser (e.g., the Russian one) leads to better performance of SCOI. Details of the relation between parser and SCOI's performance can be found in Appendix~\ref{app:spacy-cap}. On top of that, other factors like the nature of different languages' syntax might also contribute to the fluctuations across languages, which we leave for future work.

XGLM's zero-shot COMET scores on "out of English" directions are negative values. This might be due to that XGLM fails to follow the machine translation task in the prompt and sometimes produces a wrong language.

We also experiment on GPT-3.5~\cite{ouyang2022training}, which is an API-based LLM. Results are presented in Appendix~\ref{app:gpt}.

\subsection{Ablation Study}

\begin{table}[htbp]
\scalebox{0.68}{
  \centering
    \begin{tabular}{cccccccc}
    \toprule
    \multirow{2}[4]{*}{\textbf{Method}} & \multicolumn{3}{c}{\textbf{Into EN}} & \multicolumn{3}{c}{\textbf{Out of EN}} & \multirow{2}[4]{*}{\textbf{Avg.}} \\
\cmidrule(lr){2-4}\cmidrule(lr){5-7}          & \textbf{DE} & \textbf{FR} & \textbf{RU} & \textbf{DE} & \textbf{FR} & \textbf{RU} &  \\
    \midrule
    SCOI  & \textbf{64.67} & 71.26 & \textbf{54.08} & \textbf{44.87} & 55.31 & 46.47 & \textbf{56.11} \\
    \midrule
    w/o syntax & 64.44 & \textbf{71.52} & 53.33 & 43.99 & 55.52 & 46.22 & 55.84 \\
    w/o word & 63.84 & 70.95 & 53.30 & 42.55 & \textbf{56.25} & \textbf{46.82} & 55.62 \\
    \bottomrule
    \end{tabular}}
  \caption{Ablation results of SCOI on XGLM. "w/o syntax" refers to select using word-level coverage only and "w/o word" refers to select using syntax-level coverage only.}
  \label{tab:ablation}
\end{table}

To explore the effect of syntactic and lexical information, we perform ablation experiments using XGLM. Since SCOI uses both syntactic and lexical coverage, we evaluate the syntactic coverage-only and lexical coverage-only selection methods.

As shown in Table~\ref{tab:ablation}, either word-only or syntax-only coverage has limitations on some translation directions. For instance, the syntax-coverage method performs poorly on "FR-EN" and "EN-DE" while the word-coverage one performs less competitively on "RU-EN" and "EN-DE". With the help of alternate word-coverage and syntax-coverage, our proposed method of combined coverage makes the best of both worlds by and large, performing satisfactorily on all directions except "EN-FR" and achieves the highest average score.

\subsection{Experiments with Different Selection Modes}
\begin{table}[htbp]
\scalebox{0.73}{
  \centering
    \begin{tabular}{cccccccc}
    \toprule
    \multirow{2}[4]{*}{\textbf{Mode}} & \multicolumn{3}{c}{\textbf{Into EN}} & \multicolumn{3}{c}{\textbf{Out of EN}} & \multirow{2}[4]{*}{\textbf{Avg.}} \\
\cmidrule(lr){2-4}\cmidrule(lr){5-7}          & \textbf{DE} & \textbf{FR} & \textbf{RU} & \textbf{DE} & \textbf{FR} & \textbf{RU} &  \\
    \midrule
    BM25  & 63.21 & \textbf{71.36} & 52.48 & 44.13 & 55.54 & 44.58 & 55.22 \\
    \midrule
    Top-$k$ & 64.15 & 70.79 & 53.71 & 43.22 & 54.75 & \textbf{46.49} & 55.52 \\
    DPP   & 63.64 & 70.71 & 53.65 & 43.61 & \textbf{55.55} & 45.48 & 55.44 \\
    SCOI  & \textbf{64.67} & 71.26 & \textbf{54.08} & \textbf{44.87} & 55.31 & 46.47 & \textbf{56.11} \\
    \bottomrule
    \end{tabular}}
  \caption{COMET scores of 4-shot ICL performance on XGLM of different selection modes, all trying to make use of syntactic information.}
  \label{tab:mode}
\end{table}

We explore different modes of in-context example selection including top-$k$, DPP and our proposed coverage-based SCOI using XGLM to see how to make the most of syntactic information in in-context example selection. 

\paragraph{Top-$k$}
We select the top-$k$ examples with the highest syntactic similarity based on the polynomial distance used in \citet{liu2022quantifying} for each test input from the example database. Note that we can write polynomial terms as term vectors as shown in Equation~\ref{eq:vec}. In this way, a polynomial $P$ can be written as a set of term vectors $\mathcal{V}_P$. Then, following~\citet{liu2022quantifying}, we compute the distance between two polynomials ($P$ and $Q$) as:
\begin{equation}
\label{eq:polydistance}
\small
d(P, Q) = \frac{\sum\limits_{s \in \mathcal{V}_P}{\min\limits_{t \in \mathcal{V}_Q} {\parallel s - t \parallel}_1} + \sum\limits_{t \in \mathcal{V}_Q}{\min\limits_{s \in \mathcal{V}_P} {\parallel s - t \parallel}_1}}{\mid \mathcal{V}_P \mid + \mid \mathcal{V}_Q \mid},
\end{equation}
where ${\parallel s - t \parallel}_1$ is the Manhattan distance \citep{Craw2017} between term vector $s$ and $t$. Instances with top-$k$ lowest polynomial distances to the test input are used as the in-context examples.

\paragraph{DPP}
\label{app:desc_dpp}
Inspired by \citet{pmlr-v202-ye23c} and \citet{yang2023representative}, we explore selecting in-context examples for MT using Determinantal Point Processes (DPPs). DPPs are elegant probabilistic models capable of selecting a representative subset from a larger, potentially redundant set.

To incorporate both lexical diversity (differences in vocabulary coverage between different examples) and syntactic relevance (similarity between the candidate example and the test input) in the in-context example selection process, we utilize the same equation that combines diversity and relevance as used in \citet{pmlr-v202-ye23c}:
\begin{equation}
\label{eq:dpp}
    \log \det (\mathbf{L'}_S) = \frac{1}{\lambda} \sum_{i \in S} r_i + \log \det (\mathbf{L}_S),
\end{equation}
where $r_i$ represents syntactic relevance, measured by the normalized polynomial distance between each candidate example and the test input, and $\mathbf{L}_S$ denotes lexical diversity, constructed through the dot product of word vectors of all candidate examples.

Given the $\log \det (\mathbf{L'}_S)$, we can select the representative subset $S_\text{best}$ of size $k$ as follows:
\begin{equation}
    S_\text{best} = \operatornamewithlimits{argmax}\limits_{S \subseteq Z, |S| = k} \det(\mathbf{L'}_S).
\end{equation}

For the actual selection of $S_\text{best}$, we utilize the exact implementation of the greedy algorithm from \citet{pmlr-v202-ye23c}, originally proposed in \citet{fast-greedy-map}. Other details of DPP are presented in Appendix~\ref{app:dpp}.

\paragraph{Results} Results are shown in Table~\ref{tab:mode}. Note that all the methods re-rank on the basis of top-100 examples of each test input retrieved by BM25. Thus BM25 is a comparable baseline.

The top-$k$ mode does achieve a slightly higher average score compared with BM25 but in fact shows some performance drop on "FR-EN", "EN-DE" and "EN-FR" directions. This indicates simply re-ranking based on only syntactic closeness cannot necessarily secure improvement.

The DPP mode shows a slight improvement on average, but its performance fluctuates across translation directions. This indicates that simply incorporating syntax similarity into the relevance term in DPP does not necessarily yield desired improvement and how to effectively combine lexical and syntactic information using DPPs still requires exploration, which we leave for future work.

SCOI, however, performs better compared with the baselines above, obtaining highest or competitive scores across all 6 translation directions and getting the highest average score. This proves that selecting examples based on syntactic and lexical coverage alternately effectively leverages syntactic information and leads to better ICL performance.

\begin{table}[htbp]
\scalebox{0.67}{
  \centering
    \begin{tabular}{cccccccc}
    \toprule
    \multirow{2}[4]{*}{\textbf{Order}} & \multicolumn{3}{c}{\textbf{Into EN}} & \multicolumn{3}{c}{\textbf{Out of EN}} & \multirow{2}[4]{*}{\textbf{Avg.}} \\
\cmidrule(lr){2-4}\cmidrule(lr){5-7}          & \textbf{DE} & \textbf{FR} & \textbf{RU} & \textbf{DE} & \textbf{FR} & \textbf{RU} &  \\
    \midrule
    Word First & 64.45 & 70.64 & 53.76 & \textbf{45.39} & \textbf{55.91} & 45.63 & 55.96 \\
    Syntax First & \textbf{64.67} & \textbf{71.26} & \textbf{54.08} & 44.87 & 55.31 & \textbf{46.47} & \textbf{56.11} \\
    \bottomrule
    \end{tabular}}
  \caption{COMET scores of 4-shot ICL performance on XGLM of different orders of alternating.}
  \label{tab:order}
\end{table}

\subsection{Analysis on the Selection Order}
In this section, we analyze the effect of the order of alternating during the selection of SCOI.

By default, the order of alternating is syntax-first, i.e., selecting odd-numbered examples using syntactic coverage and even-numbered ones using lexical coverage. We experiment on a reversed order (i.e., word-first) for comparison.

Experimental results on XGLM are shown in Table~\ref{tab:order}. On average, the syntax-first order is slightly better than the word-first one. This indicates that focusing on syntax first can organize a better set of in-context examples.

\subsection{Analysis on the Measure of Coverage}
\label{sec:sim}
\begin{table}[htbp]
\scalebox{0.6}{
  \centering
    \begin{tabular}{cccccccc}
    \toprule
    \multirow{2}[4]{*}{\textbf{Coverage}} & \multicolumn{3}{c}{\textbf{Into EN}} & \multicolumn{3}{c}{\textbf{Out of EN}} & \multirow{2}[4]{*}{\textbf{Avg.}} \\
\cmidrule{2-7}          & \textbf{DE} & \textbf{FR} & \textbf{RU} & \textbf{DE} & \textbf{FR} & \textbf{RU} &  \\
    \midrule
    Cosine Similarity & 64.35 & \textbf{71.54} & 53.89 & \textbf{45.41} & \textbf{55.36} & 46.06 & 56.10 \\
    Normalized Distance & \textbf{64.67} & 71.26 & \textbf{54.08} & 44.87 & 55.31 & \textbf{46.47} & \textbf{56.11} \\
    \bottomrule
    \end{tabular}}
  \caption{COMET scores of 4-shot ICL performance on XGLM of different measures of coverage.}
  \label{tab:sim}
\end{table}

As described in Section~\ref{subsec:syncov}, we compute the coverage of polynomial terms $\mathtt{c}(s, t)$ in Equation~\ref{eq:synsetcov} by Equation~\ref{eq:manhattan} and \ref{eq:norm}, which is the normalized Manhattan distance between two term vectors. For comparison, we also explore cosine similarity as the measure of coverage:
\begin{equation}
    \mathtt{c}(s, t) = \frac{v_s \cdot v_t}{\parallel v_s \parallel \parallel v_t \parallel},
\end{equation}
where $v_s$ and $v_t$ are the vectors described in Equation~\ref{eq:vec} representing terms $s$ and $t$ respectively. Thus, $\mathtt{c}(s, t)$ is measured by the cosine similarity between $v_s$ and $v_t$.

Experimental results are shown in Table~\ref{tab:sim}. The difference of performance between the two measures is not significant and thus we infer that the measure of coverage has little effect on the performance of SCOI.

\subsection{Case Analysis}
\label{sec:case}
\begin{table*}[htbp]
\scriptsize
  \centering
    \begin{tabular}{m{1.7cm}<{\centering}m{6.5cm}m{6.5cm}}
    \bottomrule
     & \textbf{DE} & \textbf{EN} \\
    \hline
    Input \& Gold & Nach einer Woche voller Verluste in der Zwischenwahl erzählte Bush dem Publikum von \textbf{der Ausweitung des Handels in Asien}. & After a week of losses in the midterm election, Bush told an audience about \textbf{the expansion of trade in Asia}. \\
    \hline
    BM25 Prediction & \centering{-} & After a week of losses in the mid-election campaign, President Bush told his audience \textbf{that trade in Asia had been expanded}. \\
    \hline
    \hline
    Example-1 & Deshalb geht meiner Ansicht nach der Verlust von Sprachen mit dem Verlust von Lebensweisen einher. & I think, therefore, that if we lose languages we lose forms of life. \\
    \hline
    Example-2 & Ich stimme mit dem Standpunkt der Berichterstatterin überein und bin mit den eingeführten Veränderungen, wie der Ausweitung der Mindestdauer des Mutterschaftsurlaubs von 14 auf 20 Wochen, dem Grundsatz einer Bezahlung in voller Höhe des bisherigen Einkommens, der Einführung von Gesundheitsschutzbestimmungen am Arbeitsplatz und dem Verbot der Kündigung, einverstanden. & I agree with the position of the rapporteur and with the changes introduced, such as the extension of the minimum period for maternity leave from 14 to 20 weeks, the principle of pay equivalent to complete earnings, the establishment of health and safety requirements in the workplace, and the prohibition of dismissal. \\
    \hline
    Example-3 & Es muss eine grundlegende Strategie sein, die alle Ursachen der Krise einbezieht: die Veränderung der Ernährungsgewohnheiten in Asien, die rasche Ausweitung des Anbaus von Biokraftstoffen usw. & It must be a basic strategy that tackles all the causes of the crisis: changing food habits in Asia, the rapid rise in the cultivation of biofuels, etc. \\
    \hline
    Example-4 & Das hat seinen Widerhall bei seinem Publikum gefunden, von dem in dieser Woche 50.000 die Online-Petition für seine Freilassung unterzeichnet haben. & This has resonated among his audience, 50 000 of whom have this week signed the online petition asking for his release. \\
    \hline
    SCOI Prediction & \centering{-} & After a week of losses in the mid-term election, Bush told the audience about \textbf{the expansion of trade in Asia}. \\
    \toprule
    \end{tabular}
  \caption{An end-to-end German-to-English translation example. "Input \& Gold" refers to the test input and the gold reference. "BM25 Prediction" refers to XGLM's prediction given the test input and examples selected by BM25, which are shown in Appendix~\ref{app:case}. "Example-$i$" refers to the $i$-th example selected by SCOI. "SCOI Prediction" shows the predict of XGLM given the test input and the 4 in-context examples selected by SCOI.}
  \label{tab:caseour}
\end{table*}

An end-to-end German-to-English case is presented in Table~\ref{tab:caseour}, showing the test input, ground truth, selected examples of SCOI and model prediction of XGLM with in-context examples selected by BM25 and SCOI separately. The set of in-context examples selected by SCOI brings a good demonstration at both syntax level and word level. For instance, the first example, which is selected based on syntactic coverage, shows very close syntactic structure to the test input, with multiple prepositional phrases ("meiner Ansicht nach", "von Sprachen", "mit dem Verlust von Lebensweisen"), a very alike structure of main clause (a verb and a noun phrase) and similarly complex noun phrases ("der Verlust von Sprachen mit dem Verlust von Lebensweisen"). The second example, which is selected based on lexical coverage, covers many words as expected ("einer", "voller", "Ausweitung", etc.). The third example, again selected based on syntactic coverage, again shows very homologous syntactic structure including use of multiple prepositional phrases, complex noun phrases and similar main clause. The fourth example, based on lexical coverage, covers some other important words ("Publikum", "Woche", etc).

Table~\ref{tab:caseour} also compares SCOI's system output with that of BM25. BM25 fails to construct the proper syntactic structure when translating the German phrase "der Ausweitung des Handels in Asien" and turns it into a reported clause "that trade in Asia had been expanded", thus losing accuracy. Note that "der Ausweitung des Handels in Asien" (the expansion of trade in Asia) does not include temporal information and it could be a bygone, a current state or a future trend, while the result of BM25 assumes that it is something that happened in the past, which is inconsistent with the original meaning of the input sentence. However, SCOI, combining syntactic and lexical coverage, is able to output the exact noun phrase "the expansion of trade in Asia", which is consistent with the syntactic structure in the source German sentence and much more accurate in translation. For the complete end-to-end case of BM25, please refer to Appendix~\ref{app:case}.

\section{Conclusion}
In this work, we introduce syntactic information to in-context example selection for MT. First, we measure set-level syntactic coverage with coverage of polynomial terms based on a simplified algorithm that converts syntax trees into polynomials. Then, we propose to select in-context examples for MT based on syntactic and lexical coverage alternately to combine information of syntax and word. Our proposed method obtains the highest average COMET score among all learning-free methods, indicating that combining syntactic and lexical coverage during in-context example selection is helpful for MT. We call on the NLP community to pay more attention to syntactic knowledge for syntax-rich tasks like MT.

\section*{Acknowledgments}
This work is supported by the National Natural Science Foundation of China (62076008) and the Key Project of Natural Science Foundation of China (61936012).

\section*{Limitations}
\paragraph{Syntax Parser}
Our syntax-based method is based on reliable parsers and might not work well for low-resource languages. Meanwhile, dependency parsing could be costly when dealing with large datasets, which makes SCOI more time-consuming in such situations.

\paragraph{Semantics}
We have not tried semantic information (e.g., sentence embeddings) in our method.

\paragraph{Word-level Coverage}
We have not tried other advanced word-level coverage methods (e.g., weighted words based on their frequencies or n-gram features).

\paragraph{The Original Tree-to-Polynomial Algorithm}
Due to limited time, we have not completed the evaluation of the original tree-to-polynomial algorithm on our method to compare with our simplified version. In fact, the algorithm got stuck at a long sentence with a large dependency tree and failed to finish that instance before we killed the process due to overlong running time.

\paragraph{The Simplified Tree-to-Polynomial Algorithm}
There might be some information loss in the simplified tree-to-polynomial algorithm. For example, each term in the polynomial only presents the number of each dependency label on its corresponding root-to-node path but cannot show the exact order of these labels. In other words, our simplified tree-to-polynomial algorithm is a many-to-one mapping and is thus irreversible.

\section*{Ethics Statement}
\paragraph{Computational Budget}
\begin{table}[htbp]
\small
  \centering
    \begin{tabular}{cc}
    \toprule
    \textbf{Task} & \textbf{Time (min)} \\
    \midrule
    BM25 Pre-selection & 12 \\
    Dependency Parsing & 60 \\
    Tokenization & 4 \\
    Combined Coverage & 9 \\
    LLM Inference & 90 \\
    \bottomrule
    \end{tabular}
  \caption{Average computation time on German into/out of English using XGLM.}
  \label{tab:time}
\end{table}
We run pre-processing and in-context example selection on Intel$^\circledR$ Xeon$^\circledR$ Gold 6348 CPU and the LLM's inference on NVIDIA A40 (we set batch size to 2). Table \ref{tab:time} presents the average computation time, with XGLM as the LLM. The major bottleneck of computation time lies in syntax parsing, which is due to the large size of the example database.

\paragraph{Reproducibility}
All the experiments are reproducible since all the methods are deterministic and sampling is disabled during LLM generation.

\paragraph{Scientific Artifacts}
\begin{table}[htbp]
\small
  \centering
    \begin{tabular}{ll}
    \toprule
    \textbf{Artifact} & \textbf{License} \\
    \midrule
    spaCy & MIT \\
    Sacremoses & MIT \\
    retriv & MIT \\
    XGLM  & MIT \\
    Alpaca & Apache-2.0 \\
    COMET & Apache-2.0 \\
    sacreBLEU & Apache-2.0 \\
    FLORES-101 & CC-BY-SA-4.0 \\
    Europarl & Unknown \\
    ParaCrawl & CC0 \\
    CTQ Scorer & MIT \\
    \bottomrule
    \end{tabular}%
  \caption{Licenses of scientific artifacts we use.}
  \label{tab:artifact}%
\end{table}%

We cite all the creators of scientific artifacts we use in this paper. Licenses of these scientific artifacts are shown in Table \ref{tab:artifact}. Our use of these artifacts is consistent with their intended use.

\normalem
\bibliography{anthology,custom}

\appendix

\section{Analysis of Tree-to-Polynomial Algorithms}
\label{app:analysis}
\subsection{Original Algorithm from \citet{liu2022quantifying}}
\label{subapp:original}

We denote the cost of the algorithm if the tree has $n$ nodes by $T(n)$ (then $T(1) = O(1)$), the number of nodes in the tree rooted in node $m$ by $\mid m \mid$, the number of terms in the polynomial of node $m$ by $\parallel m \parallel$. Note that if $\mid m \mid = n$, then 
\begin{equation}
\label{eq:sum-of-n}
\sum_{i=1}^k{\mid n_i \mid} = n-1.
\end{equation}

For simplicity, we assume that the cost of addition of polynomial terms is the same as that of multiplication.

To get the polynomial of $m^l$ in Equation~\ref{eq:original}, we need to compute the polynomial of each $n_i$ (each is $T(\mid n_i \mid)$ and the sum is $T_1(n) = \sum_{i=1}^k{T(\mid n_i \mid)}$) and the multiplication of the former polynomials (which is the sum of the multiplication of all possible combinations of terms from the child nodes and each combination requires multiplying $k$ terms together plus an addition thus the overall cost should be $ T_2(n) = O((1 + (k-1))\cdot \prod_{i=1}^k{\parallel n_i \parallel})$)~\footnote{Here the additions include the addition of the whole product of former polynomials and $y_l$.}. Then the overall cost of computing Equation~\ref{eq:original} is

\begin{multline}
\label{eq:ori-analaysis}
    T(n) = T_1(n) + T_2(n)\\
    = \sum_{i=1}^k{T(\mid n_i \mid)} + O(k\cdot \prod_{i=1}^k{\parallel n_i \parallel}).
\end{multline}

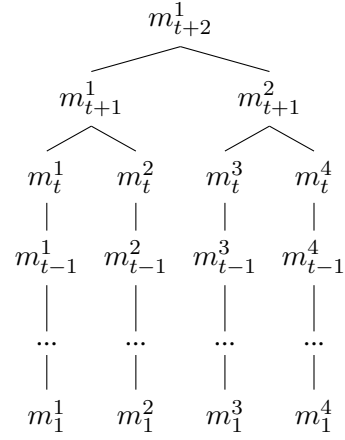
\begin{figure}
    \centering
    \begin{tikzpicture}
    \Tree
    [.$m^1_{t+2}$
    [.$m^1_{t+1}$
    [.$m^1_{t}$ [.$m^1_{t-1}$ [.{...} [.$m^1_1$ ] ] ] ] 
    [.$m^2_{t}$ [.$m^2_{t-1}$ [.{...} [.$m^2_1$ ] ] ] ]
    ] [.$m^2_{t+1}$
    [.$m^3_{t}$ [.$m^3_{t-1}$ [.{...} [.$m^3_1$ ] ] ] ] 
    [.$m^4_{t}$ [.$m^4_{t-1}$ [.{...} [.$m^4_1$ ] ] ] ]
    ] ]
    \end{tikzpicture}
    \caption{An example tree with $t+2$ layers and $4t+3$ nodes. $m^j_i$ denotes the $j$-th node on the $i$-th layer.}
    \label{fig:analysistree}
\end{figure}

Consider a tree as shown in Figure~\ref{fig:analysistree} with $t+2$ layers and $4t+3$ nodes, where $m^j_i$ denotes the $j$-th node on the $i$-th layer. The cost of computing the polynomial of $m^j_t$ should be $O(t)$ since it is just to add $t$ single-variable terms together according to Equation~\ref{eq:original}. Then, the cost of computing the polynomial of $m^i_{t+1}$ should be $O(2t + 2t^2)$ according to Equation~\ref{eq:ori-analaysis}, which can be further simplified to $O(t^2)$. Finally, the cost of computing the polynomial of $m^1_{t+2}$, which is also the polynomial representing the whole tree, should be $O(2t^2 + 2(t^2)^2)$ according to Equation~\ref{eq:ori-analaysis}, which can be further simplified to $O(t^4)$. Thus in this tree, the cost is:
\begin{equation}
\label{eq:t-analysis}
    \hat{T}(4t+3) = O(t^4).
\end{equation}
Let $s = 4t+3$, we simplify Equation~\ref{eq:t-analysis} and have:
\begin{equation}
    \hat{T}(s) = O(s^4).
\end{equation}
Therefore, we prove that the original tree-to-polynomial algorithm can be of quartic time complexity in some cases.

In fact, given any constant $p=2^q$ where $q$ is a positive integer, we can construct a tree in the way as shown in Figure~\ref{fig:analysistree} with $t + q$ layers and $pt + p - 1$ nodes. Let $m^1_{t+q}$ denote the root node. For each $i$ between $1$ and $q$ and each $j$ between $1$ and $2^{q-i}$, $m^j_{t+i}$ has two child nodes $m^{2j-1}_{t+i-1}$ and $m^{2j}_{t+i-1}$. For each $i$ between $2$ and $t$ and each $j$ between $1$ and $2^q$, $m^j_i$ has only one child node $m^j_{i-1}$. Finally, for each $j$ between $1$ and $2^q$, $m^j_1$ is the leaf node. In this way, the cost of computing the polynomial of $m^j_t$ is $O(t)$ as discussed above. That of $m^j_{t+1}$, $m^j_{t+2}$, ..., $m^j_{t+q}$ should be $O(t^2)$, $O(t^4)$, ..., $O(t^{2^q})$ recursively, the last one with $q$ times of recursion being $\hat{T}(pt+p-1)$ indeed. Let $s = pt+p-1$, we again ignore the constant factors and insignificant terms and then have:
\begin{equation}    
\hat{T}(s)=O(s^{2^q})=O(s^p).
\end{equation}
Thus we prove that the complexity of the original tree-to-polynomial algorithm can be polynomial of arbitrarily large degree $p=2^q$ in some cases. So when dealing with very large dependency trees of long sentences, the original algorithm can be quite time-consuming and thus impractical for application in MT where there can be millions of data to be processed.

However, we have not proven the exact lower bound of the cost or the average cost according to Equation~\ref{eq:ori-analaysis}, which we leave for future work.

\subsection{Our Simplified Algorithm}
We use the same symbols as in Section~\ref{subapp:original}. Given $\mid m \mid = n$, Equation~\ref{eq:sum-of-n} still holds in this section. Moreover, in our simplified algorithm, the number of terms in a polynomial equals to the number of nodes in the tree rooted in the corresponding node:
\begin{equation}
    \parallel m \parallel = \mid m \mid,
\end{equation}
and
\begin{equation}
\label{eq:num-of-terms}
    \parallel n_i \parallel = \mid n_i \mid.
\end{equation}

To get the polynomial of $m^l$ in Equation~\ref{eq:simplified}, we need to compute the polynomial of each $n_i$ (each is $T(\mid n_i \mid)$), the sum of $1$ and all the former polynomials (which is $O(\sum_{i=1}^k{\parallel n_i \parallel})$), the multiplication of $x_l$ (which can be seen as multiply $x_l$ with all the terms in the former polynomials plus $1$ and thus should be $O(1+\sum_{i=1}^k{\parallel n_i \parallel})$ and can be further simplified to $O(\sum_{i=1}^k{\parallel n_i \parallel})$). Then the overall cost of computing Equation~\ref{eq:simplified} is
\begin{equation}
    T(n) = \sum_{i=1}^{k}T(\mid n_i \mid) + 2\cdot O(\sum_{i=1}^k{\parallel n_i \parallel}).
\end{equation}
We then apply Equation~\ref{eq:sum-of-n} and \ref{eq:num-of-terms} and ignore the constant factors to get
\begin{equation}
    T(n) = \sum_{i=1}^{k}T(\mid n_i \mid) + O(n).
\end{equation}
Then
\begin{equation}
\label{eq:rec-first}
    T(n) - \sum_{i=1}^{k}T(\mid n_i \mid) = O(n).
\end{equation}
Analogously,
\begin{equation}
    \forall 1\leq i \leq k, T(\mid n_i\mid) - \sum_{j=1}^{k_i}{T(\mid n_{i_j} \mid)} = O(\mid n_i\mid),
\end{equation}
where $n_i$ has $k_i$ child nodes denoted by $n_{i_j}$. Thus
\begin{equation}
\label{eq:rec-second}
\begin{split}
    \sum_{i=1}^{k}T(\mid n_i \mid) - \sum_{i=1}^{k}{\sum_{j=1}^{k_j}{T(\mid n_{i_j} \mid)}} = \sum_{i=1}^{k}{O(\mid n_i \mid)}\\ = O(n-1).
\end{split}
\end{equation}
With the recursive boundary
\begin{equation}
    T(1) - 0 = O(1),
\end{equation}
we can continue the process recursively (in fact, each level of recursion corresponds to a layer in the tree) until each node has appeared on left-hand side and add together Equation~\ref{eq:rec-first}, \ref{eq:rec-second} and so on to get
\begin{equation}
\begin{split}
    T(n) = O(n) + O(n-1) + O(n-1-k) + ...\\
    \leq O(n^2).
\end{split}
\end{equation}
Thus we prove that the complexity of our simplified tree-to-polynomial algorithm is no more than quadratic time.

\section{BLEU Results}
\label{app:bleu}
\begin{table*}[htbp]
\small
  \centering
    \begin{tabular}{cccccccc}
    \toprule
    \multirow{2}[4]{*}{\textbf{System}} & \multicolumn{3}{c}{\textbf{Into EN}} & \multicolumn{3}{c}{\textbf{Out of EN}} & \multirow{2}[4]{*}{\textbf{Avg.}} \\
\cmidrule(lr){2-4}\cmidrule(lr){5-7}          & \textbf{DE} & \textbf{FR} & \textbf{RU} & \textbf{DE} & \textbf{FR} & \textbf{RU} &  \\
    \midrule
    \midrule
    \multicolumn{8}{c}{\textbf{XGLM}} \\
    \midrule
    Zero-shot & 31.13 & 32.68 & 23.96 & 10.41 & 17.8 & 5.56 & 20.26 \\
    \midrule
    \multicolumn{8}{l}{\textit{Learning-free}} \\
    Random & 31.31 & 32.68 & \textbf{24.85} & 19.63 & 28.79 & 17.57 & 25.81 \\
    BM25  & 31.06 & \textbf{33.34} & 24.47 & 20.16 & \textbf{29.79} & 18.18 & 26.17 \\
    R-BM25 & 31.16 & 32.99 & 24.71 & 20.00 & 29.17 & 17.93 & 25.99 \\
    Fuzzy & \textbf{31.95} & 33.08 & 24.42 & 20.29 & 29.77 & 18.01 & \textbf{26.25} \\
    SCOI (\textit{ours}) & 31.51 & 32.88 & \textbf{24.85} & \textbf{20.45} & 29.39 & \textbf{18.25} & 26.22 \\
    \midrule
    \multicolumn{8}{l}{\textit{Learning-based}} \\
    CTQ Scorer & 32.02 & 32.35 & 25.29 & 20.94 & 30.59 & 18.53 & 26.62 \\
    \midrule
    \midrule
    \multicolumn{8}{c}{\textbf{Alpaca}} \\
    \midrule
    Zero-shot & 33.57 & 35.73 & 26.25 & 20.86 & 29.08 & 15.55 & 26.84 \\
    \midrule
    \multicolumn{8}{l}{\textit{Learning-free}} \\
    Random & 33.50 & \textbf{36.24} & 26.48 & 20.08 & 29.05 & 15.82 & 26.86 \\
    BM25  & 33.16 & 35.11 & 26.58 & 20.23 & \textbf{29.76} & 15.99 & 26.81 \\
    R-BM25 & 33.47 & 35.42 & 26.23 & 20.46 & 29.64 & 16.27 & 26.92 \\
    Fuzzy & 33.51 & 35.51 & 26.02 & 20.26 & 29.58 & 15.87 & 26.79 \\
    SCOI (\textit{ours}) & \textbf{33.93} & 35.44 & \textbf{26.69} & \textbf{20.70} & 29.61 & \textbf{16.53} & \textbf{27.15} \\
    \midrule
    \multicolumn{8}{l}{\textit{Learning-based}} \\
    CTQ Scorer & 33.75 & 35.83 & 26.56 & 20.99 & 30.23 & 16.26 & 27.27 \\
    \bottomrule
    \end{tabular}
  \caption{BLEU scores of 4-shot ICL performance of SCOI and other methods for translation on all 6 directions of 2 language models. The zero-shot baseline of each model is listed in the first row. All the methods except CTQ Scorer are learning-free, which do not require task, language or LLM-specific training. "\textbf{Avg.}" refers to the average score across all 6 directions. The highest scores among learning-free methods are in \textbf{bold} text.}
  \label{tab:main-bleu}
\end{table*}
The BLEU scores of our main results are shown in Table~\ref{tab:main-bleu}.

\section{The spaCy Models Used for Parsing}
\label{app:spacy}
\begin{table}[htbp]
  \centering
    \begin{tabular}{ccc}
    \toprule
    \textbf{Language} & \textbf{spaCy Model} & \textbf{Version} \\
    \midrule
    DE    & \texttt{de\_core\_news\_sm} & 3.7.0 \\
    EN    & \texttt{en\_core\_web\_sm} & 3.7.1 \\
    FR    & \texttt{fr\_core\_news\_sm} & 3.7.0 \\
    RU    & \texttt{ru\_core\_news\_sm} & 3.7.0 \\
    \bottomrule
    \end{tabular}%
  \caption{The spaCy models and their versions of different languages used for dependency parsing.}
  \label{tab:spacy}
\end{table}

The spaCy models and their corresponding versions we use for dependency parsing are listed in Table~\ref{tab:spacy}.


\section{Effect of Parser}
\label{app:spacy-cap}
\begin{table}[htbp]
\scalebox{0.8}{
  \centering
    \begin{tabular}{cccc}
    \toprule
    \textbf{Direction} & \textbf{$\Delta$} & \textbf{Parser} & \textbf{LAS} \\
    \midrule
    DE-EN & +1.46 & \texttt{de\_core\_news\_sm} & 0.90 \\
    FR-EN & -0.10 & \texttt{fr\_core\_news\_sm} & 0.83 \\
    RU-EN & +1.60 & \texttt{ru\_core\_news\_sm} & 0.95 \\
    Out of EN (Avg.) & +0.80 & \texttt{en\_core\_web\_sm} & 0.90 \\
    \bottomrule
    \end{tabular}}
  \caption{Performance gains ("$\Delta$") of SCOI over BM25 using XGLM and capabilities of corresponding parsers on different translation directions. "LAS" refers to the labeled attachment score of a parser.}
  \label{tab:spacy-cap}
\end{table}

In order to better understand the relation between the performance of SCOI and the capability of the parser, we compare the labeled attachment scores (LAS) of different parsers used in our experiments reported on the official website of spaCy~\footnote{https://spacy.io/models/}. Table~\ref{tab:spacy-cap} shows performance gains of SCOI over the BM25 baseline using XGLM and capabilities of
corresponding parsers. The results show that a better parser leads to better performance of SCOI and incidate that SCOI is highly dependent on parsers.

\section{Results on GPT-3.5}
\label{app:gpt}
\begin{table*}[htbp]
\small
  \centering
    \begin{tabular}{cccccccc}
    \toprule
    \multirow{2}[4]{*}{\textbf{System}} & \multicolumn{3}{c}{\textbf{Into EN}} & \multicolumn{3}{c}{\textbf{Out of EN}} & \multirow{2}[4]{*}{\textbf{Avg.}} \\
\cmidrule(lr){2-4}\cmidrule(lr){5-7}          & \textbf{DE} & \textbf{FR} & \textbf{RU} & \textbf{DE} & \textbf{FR} & \textbf{RU} &  \\
    \midrule
    \multicolumn{8}{l}{\textit{Learning-free}} \\
    Random & 77.52 & 81.78 & \textbf{67.02} & 69.04 & 84.12 & 71.26 & 75.12 \\
    BM25  & \textbf{77.54} & 81.60 & 66.17 & 68.93 & 84.06 & 71.73 & 75.01 \\
    R-BM25 & 77.24 & 81.54 & 66.45 & \textbf{69.25} & 84.08 & 71.46 & 75.00 \\
    Fuzzy & 77.36 & \textbf{81.89} & 66.52 & 68.83 & \textbf{84.33} & \textbf{72.49} & \textbf{75.24} \\
    SCOI (\textit{ours})  & 77.17 & \textbf{81.89} & 66.38 & 69.07 & 84.31 & 72.13 & 75.16 \\
    \midrule
    \multicolumn{8}{l}{\textit{Learning-based}} \\
    CTQScorer & 77.40 & 81.99 & 66.77 & 69.33 & 83.78 & 73.06 & 75.39 \\
    \bottomrule
    \end{tabular}
  \caption{Results on GPT-3.5.}
  \label{tab:gpt}
\end{table*}

We call OpenAI's API~\footnote{https://openai.com/api/} of \texttt{gpt-3.5-turbo-0125} to evaluate different in-context example selection methods on GPT-3.5. Results are presented in Table~\ref{tab:gpt}.

It seems the difference between in-context example selection methods is not so significant as that on smaller LLMs. This might be because that the capability of GPT-3.5 has been strong enough so that in-context examples bring limited help. For such large-scaled models, design and organization of prompt and use of additional information or knowledge might be more crucial in improving performance of ICL.



\section{Details of DPPs}
\label{app:dpp}

We set the $\lambda$ in Equation \ref{eq:dpp} to $0.5$ to balance syntactic relevance and lexical diversity. As mentioned in Section \ref{app:desc_dpp}, the word vectors $\mathbf{W}_{N \times T}$, used to compute lexical diversity, where $N$ is the number of documents (candidate examples) and $T$ is the number of terms (words) in each test input, are constructed as follows:
\begin{equation}
    \mathbf{W}_{i,j} = \text{idf}_j \times \left( \frac{\text{tf}_{i,j} \times (k_1 + 1)}{\text{tf}_{i,j} + k_1 \times (1 - b + b \times l_i)} \right),
\end{equation}

where $i$ and $j$ refer to the $i$-th candidate example and the $j$-th term in a test input, respectively. Here, $\text{idf}_j$ is the inverse document frequency of the $j$-th term across all candidate examples, $\text{tf}_{i,j}$ is the term frequency of the $j$-th term in the $i$-th candidate example, and $l_i$ is the length of the $i$-th candidate example. The parameters $k_1$ and $b$ are hyperparameters.

\section{The Example of BM25}
\label{app:case}
\begin{table*}[htbp]
\scriptsize
  \centering
    \begin{tabular}{|m{1.5cm}<{\centering}|m{6.5cm}|m{6.5cm}|}
    \hline
     & \textbf{DE} & \textbf{EN} \\
    \hline
    Input \& Gold & Nach einer Woche voller Verluste in der Zwischenwahl erzählte Bush dem Publikum von der Ausweitung des Handels in Asien. & After a week of losses in the midterm election, Bush told an audience about the expansion of trade in Asia. \\
    \hline
    Example-1 & Ich stimme mit dem Standpunkt der Berichterstatterin überein und bin mit den eingeführten Veränderungen, wie der Ausweitung der Mindestdauer des Mutterschaftsurlaubs von 14 auf 20 Wochen, dem Grundsatz einer Bezahlung in voller Höhe des bisherigen Einkommens, der Einführung von Gesundheitsschutzbestimmungen am Arbeitsplatz und dem Verbot der Kündigung, einverstanden. & I agree with the position of the rapporteur and with the changes introduced, such as the extension of the minimum period for maternity leave from 14 to 20 weeks, the principle of pay equivalent to complete earnings, the establishment of health and safety requirements in the workplace, and the prohibition of dismissal. \\
    \hline
    Example-2 & Deshalb geht meiner Ansicht nach der Verlust von Sprachen mit dem Verlust von Lebensweisen einher. & I think, therefore, that if we lose languages we lose forms of life. \\
    \hline
    Example-3 & Herr Minister, diese Woche wird von dem erklärten Willen des Europäischen Parlaments geprägt sein, gegen den Verlust der biologischen Vielfalt anzukämpfen. & Minister, this week will have been marked by the desire shown by the European Parliament to fight against the loss of biodiversity.
 \\
    \hline
    Example-4 & Nach dem, was mir erzählt wurde, nicht gut. & From what I was told I suspect they were not good. \\
    \hline
    Prediction & \centering{-} & After a week of losses in the mid-election campaign, President Bush told his audience that trade in Asia had been expanded. \\
    \hline
    \end{tabular}
  \caption{An end-to-end "DE-EN" translation example of BM25, with the same test input in Table~\ref{tab:caseour}.}
  \label{tab:casebm}
\end{table*}

An end-to-end German-to-English translation example of BM25 is shown in Table~\ref{tab:casebm}, the test input is the same as that of SCOI discussed in Section~\ref{sec:case}.

BM25 mainly focuses on lexical similarity and does not take coverage into consideration. For example, the word "Publikum" is not covered by BM25 since it is based on the Top-$k$ mode while SCOI does cover it. Moreover, it does not emphasize the similarity in syntax. Even though some examples contain similar syntactic structure (e.g., the second example is just the first example selected by our method), BM25 fails to put these examples in the front to allow LLMs pay more attention to those more helpful examples.

\end{document}